\documentclass[11pt]{article}
\usepackage{acl2016}
\usepackage{times}
\usepackage{url}
\usepackage{latexsym}
\usepackage{array}
\usepackage{multirow}
\usepackage{amssymb}
\usepackage{amsmath}
\usepackage{mathtools}
\usepackage{amsthm}
\usepackage{xspace}
\usepackage{algorithm}
\usepackage{tikz-dependency}
\usepackage{cancel}
\usepackage{float}
\usepackage{mathpartir}
\usepackage{proof}
\usepackage{subfigure}
\usepackage{graphicx}

\usepackage{tikz} 
\usepackage{tikz-qtree}
\usepackage{examples}

\aclfinalcopy 
\def\aclpaperid{521} 

\newcommand{\tuple}[1]{\ensuremath{\langle {#1} \rangle}}

\newcommand{\notes}[1]{}

 \theoremstyle{definition}
 
\theoremstyle{plain}

\newcommand{\ith}[1]{\ensuremath{i^{{th}}}}

\newcount\permx
\newcount\permy
\def\permdot#1#2{
\permx=#1 \advance\permx by-1
\permy=#2 \advance\permy by-1
\psframe[fillcolor=black, fillstyle=solid]
(\permx,\permy)(#1, #2)
}

\newcommand{\x}[1]{\ensuremath{x_{#1}}\xspace}

\newcommand{\boxnum}[1]{{\setlength{\fboxsep}{1pt}\raisebox{1pt}{\hspace{1pt}\fbox{\tiny #1}\hspace{1pt}}}}
\newcommand{\ind}[1]{\ensuremath{_{\kern-0.5pt\boxnum{#1}}}}

\def\namecite{\newcite}

\newcommand{\smallnt}[1]{\ensuremath{_{\mbox{\tiny PP}}}\xspace}

\newcommand{\leftb}{\ensuremath{\mathsf{left}}\xspace}

\newcommand{\leftc}{\ensuremath{\mathsf{left}}\xspace}
\newcommand{\rightc}{\ensuremath{\mathsf{right}}\xspace}
\newcommand{\headc}{\ensuremath{\mathsf{head}}\xspace}
\newcommand{\rootc}{\ensuremath{\mathsf{root}}\xspace}

\newcommand{\shift}{\ensuremath{\mathsf{shift}}\xspace}

\newcommand{\promote}{\ensuremath{\mathsf{pro}}\xspace}

\newcommand{\lreduce}{\ensuremath{\mathsf{re_{\small \curvearrowleft}}}\xspace}
\newcommand{\rreduce}{\ensuremath{\mathsf{re_{\small \curvearrowright}}}\xspace}
\newcommand{\ladj}{\ensuremath{\mathsf{adj_{\small \curvearrowleft}}}\xspace}
\newcommand{\radj}{\ensuremath{\mathsf{adj_{\small \curvearrowright}}}\xspace}

\newcommand{\sttop}{\ensuremath{s_0}\xspace}
\newcommand{\stnext}{\ensuremath{s_1}\xspace}

\newcommand{\pseudocode}{Algorithm}
\floatname{algorithm}{\pseudocode}

\newcommand{\olditem}[3]{\ensuremath{\tuple{{#3}, \ {#2}}}\xspace} 

\newcommand{\arcleft}[2]{\ensuremath{{#1}^\curvearrowleft{#2}}\xspace}

\iffalse

\else

\fi

\title{Incremental Parsing with Minimal Features Using Bi-Directional LSTM}

\author{James Cross \and Liang Huang\\
        School of Electrical Engineering and Computer Science\\
        Oregon State University\\
        Corvallis, Oregon, USA\\
        {\tt \{crossj,liang.huang\}@oregonstate.edu}
    }

\date{}

\begin{document}

\maketitle

\begin{abstract}

Recently, neural network approaches for parsing have largely automated the combination of individual features,  but still rely on (often a larger number of) atomic features created from human linguistic intuition, 
and potentially omitting important global context. 
To further reduce feature engineering to the bare minimum, 
we use bi-directional LSTM sentence representations to model a parser state with only three sentence positions, which automatically identifies important aspects of the entire sentence. 
This model achieves state-of-the-art results among greedy dependency parsers for English. 
We also introduce a novel transition system for constituency parsing which does not require binarization, and together with the above architecture, achieves state-of-the-art results among greedy parsers for both English and Chinese.

\end{abstract}

\section{Introduction}
\label{sec:intro}
Recently, neural network-based parsers have become popular,
with the promise of reducing the burden of manual feature engineering. 
For example, \namecite{chen2014fast} and subsequent work
replace the huge amount of 
manual feature combinations in non-neural network efforts \cite{nivre+:2006,zhang+nivre:2011}
by vector embeddings of the atomic features.
However, this approach has two related limitations.
First, it still depends on 
a large number of carefully designed atomic features. 
For example, \namecite{chen2014fast} and subsequent work such as \namecite{weiss2015google}
use 48 atomic features from \namecite{zhang+nivre:2011}, including select third-order dependencies.
More importantly, this approach inevitably leaves out some nonlocal information which could be useful.
In particular, though such a model can exploit similarities between words and other embedded categories, and learn interactions among those atomic features, it cannot exploit any other details of the text.


We aim to reduce the need for manual induction of atomic features to the bare minimum,
by using bi-directional recurrent neural networks to automatically learn context-sensitive representations 
for each word in the sentence.
This approach allows the model to learn arbitrary patterns from the entire sentence, 
effectively extending the generalization power of embedding individual words to longer sequences. 
Since such a feature representation is less dependent on earlier parser decisions, it is also more resilient to local mistakes.

With just three positional features 
we can build a greedy shift-reduce dependency parser that is on par with the most accurate parser 
in the published literature for English Treebank. This effort is similar in motivation to 
the stack-LSTM of \namecite{dyer2015transition}, but uses a much simpler architecture.

We also extend this model to predict phrase-structure trees 
with a novel shift-promote-adjoin system tailored to greedy constituency parsing,
and with just two more positional features (defining tree span) and nonterminal label embeddings
we achieve the most accurate greedy constituency parser for both English and Chinese.

\section{LSTM Position Features}

\begin{figure}[H]
\centering
\resizebox{0.45\textwidth}{3.3cm}{\resizebox{\linewidth}{!}{

\begin{tikzpicture}[shorten >=1pt,->,draw=black!50, node distance=1cm]
    \tikzstyle{every pin edge}=[<-,shorten <=1pt]
    \tikzstyle{neuron}=[circle,fill=black!25,minimum size=15pt,inner sep=0pt]
    \tikzstyle{annot} = [text width=6em, text centered];
    \tikzstyle{forward lstm}=[neuron, fill=orange!50];
    \tikzstyle{backward lstm}=[neuron, fill=blue!50];

    \foreach \name / \x in {1,...,5}
        {
        \node[annot] (O-\name) at (2*\x, -7) {$f_{\name}$;$b_{\name}$};
        \node[backward lstm] (B-\name) at (2*\x+0.5,-9) {};
        \node[forward lstm] (F-\name) at (2*\x-0.5,-10) {};
        \node[annot] (I-\name) at (2*\x, -12) {$w_{\name}$;$t_{\name}$};
        \path[orange, thick] (I-\name) edge (F-\name);
        \path[blue, thick, dashed] (I-\name) edge (B-\name);
        \path[orange, thick] (F-\name) edge (O-\name);
        \path[blue, thick, dashed] (B-\name) edge (O-\name);
        }

    \foreach \left / \right in {1/2,2/3,3/4,4/5}
        {
        \path[orange, thick, bend right] (F-\left) edge (F-\right);
        \path[blue, thick, dashed, bend right] (B-\right) edge (B-\left);
        }
\end{tikzpicture}

}}
\caption{The sentence is modeled with an LSTM in each direction whose input vectors at each time step are word and part-of-speech tag embeddings.}
\label{fig:lstm}
\end{figure}
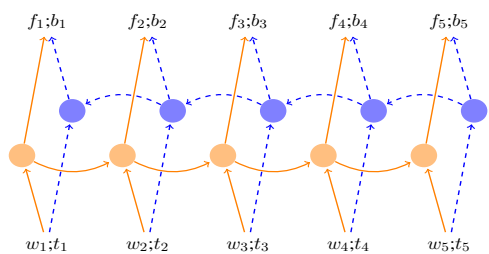

The central idea behind this approach is exploiting the power of recurrent neural networks to let the model decide what apsects of sentence context are important to making parsing decisions, rather than relying on fallible linguistic information (which moreover requires leaving out information which could be useful). In particular, we model an input sentence using Long Short-Term Memory networks (LSTM), which have made a recent resurgence after being initially formulated by \namecite{hochreiter1997long}.

The input at each time step is simply a vector representing the word, in this case an embedding for the word form and one for the part-of-speech tag. These embeddings are learned from random initialization together with other network parameters in this work. In our initial experiments, we used one LSTM layer in each direction (forward and backward), and then concatenate the output at each time step to represent that sentence position: that word in the entire context of the sentence. This network is illustrated in Figure~\ref{fig:lstm}.

\begin{figure}[H]
\centering
\resizebox{0.45\textwidth}{7cm}{\resizebox{\linewidth}{!}{

\begin{tikzpicture}[shorten >=1pt,->,draw=black!50, node distance=1cm]
    \tikzstyle{every pin edge}=[<-,shorten <=1pt]
    \tikzstyle{neuron}=[circle,fill=black!25,minimum size=15pt,inner sep=0pt]
    \tikzstyle{annot} = [text width=6em, text centered];
    \tikzstyle{forward lstm}=[neuron, fill=orange!50];
    \tikzstyle{backward lstm}=[neuron, fill=blue!50];

    \foreach \name / \x in {1,...,5}
        {
        \node[annot] (H-\name) at (2*\x, -5) {$h_{\name}$};

        \node[annot] (O2-\name) at (2*\x, -7) {$f^2_{\name}$;$b^2_{\name}$};
        \node[backward lstm] (B2-\name) at (2*\x+0.5,-9) {};
        \node[forward lstm] (F2-\name) at (2*\x-0.5,-10) {};
        \node[annot] (O1-\name) at (2*\x, -12) {$f^1_{\name}$;$b^1_{\name}$};
        \node[backward lstm] (B1-\name) at (2*\x+0.5,-14) {};
        \node[forward lstm] (F1-\name) at (2*\x-0.5,-15) {};
        \node[annot] (I-\name) at (2*\x, -17) {$w_{\name}$;$t_{\name}$};

        \path[orange, thick] (O1-\name) edge (F2-\name);
        \path[blue, thick, dashed] (O1-\name) edge (B2-\name);
        \path[orange, thick] (F2-\name) edge (O2-\name);
        \path[blue, thick, dashed] (B2-\name) edge (O2-\name);

        \path[orange, thick] (I-\name) edge (F1-\name);
        \path[blue, thick, dashed] (I-\name) edge (B1-\name);
        \path[orange, thick] (F1-\name) edge (O1-\name);
        \path[blue, thick, dashed] (B1-\name) edge (O1-\name);

        \path[red, very thick] (O2-\name) edge (H-\name);
        \path[red, very thick, bend left] (O1-\name) edge (H-\name);
        }

    \foreach \left / \right in {1/2,2/3,3/4,4/5}
        {
        \path[orange, thick, bend right] (F2-\left) edge (F2-\right);
        \path[blue, thick, dashed, bend right] (B2-\right) edge (B2-\left);

        \path[orange, thick, bend right] (F1-\left) edge (F1-\right);
        \path[blue, thick, dashed, bend right] (B1-\right) edge (B1-\left);
        }
\end{tikzpicture}

}}
\caption{In the 2-Layer architecture, the output of each LSTM layer is concatenated to create the positional feature vector.}
\label{fig:lstm2}
\end{figure}
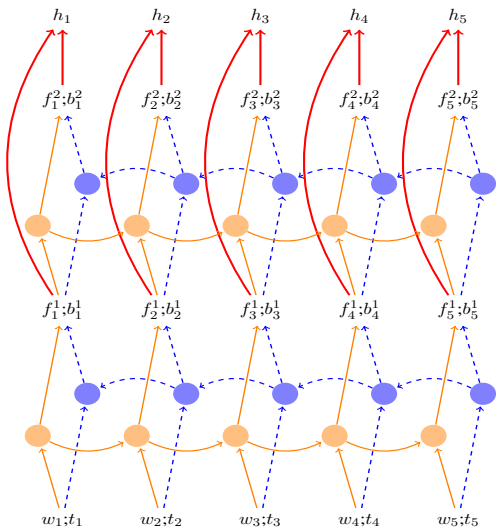

It is also common to stack multiple such LSTM layers, where the output of the forward and backward networks at one layer are concatenated to form the input to the next.  We found that parsing performance could be improved by using two bi-directional LSTM layers in this manner, and concatenating the output of both layers as the positional feature representation, which becomes the input to the fully-connected layer. This architecture is shown in Figure~\ref{fig:lstm2}.

Intuitively, this represents the sentence position by the word in the context of the sentence up to that point and the sentence after that point in the first layer, as well as modeling the ``higher-order'' interactions between parts of the sentence in the second layer.  In Section~\ref{sec:exps} we report results using only one LSTM layer (``Bi-LSTM") as well as with two layers where output from each layer is used as part of the positional feature (``2-Layer Bi-LSTM").

\section{Shift-Reduce Dependency Parsing}
\label{sec:shift-reduce}

\begin{figure}
\centering
\hspace{-0.15in}
  \begin{tabular}{ll}
   input:      & $w_0 \ldots w_{n-1}$ \\[0.06in]
    axiom        & \olditem{0}{0}{\epsilon}: $\emptyset$ \\ [0.06in]

    \small{\shift}       & \inferrule{\olditem{\ell}{j}{S}: A}{\olditem{\ell+1}{j+1}{S|j}: A} \ $j<n$ \\[0.2in]

    \small{\lreduce}     & \inferrule{\olditem{\ell}{j}{S|\stnext|\sttop}: A}
                             {\olditem{\ell+1}{j}{S|\sttop}: A\cup\{\arcleft{\stnext}{\sttop}\}} \\[0.2in]


    goal         & \olditem{2n-1}{n}{\sttop}: $A$ 
  \end{tabular}
\caption{The arc-standard
dependency parsing system \protect\cite{nivre:2008} (\rreduce omitted).
Stack $S$ is a list of heads, $j$ is the start index of the queue,
and \sttop and  \stnext are the top two head indices on $S$.
}
\label{fig:vanilla}
\end{figure}

\begin{table}
\resizebox{!}{!}{
\begin{tabular}{l|l|l}
            & {\small dependency}      & {\small constituency} \\
\hline
{\small positional}  & $s_1, s_0, q_0$ & $s_1, s_0, q_0, s_1.\leftb, s_0.\leftb$ \\ 
\hline
{\small labels}      & -               & $s_0.\{\leftc,\rightc,\rootc,\headc\}$ \\
            &                & $s_1.\{\leftc,\rightc,\rootc,\headc\}$ \\
\end{tabular}}
\caption{Feature templates. Note that, remarkably, even though we do labeled dependency parsing,
we do {\em not} include arc label as features.}
\label{tab:features}
\end{table}

We use the arc-standard system for dependency parsing (see Figure~\ref{fig:vanilla}). 
By exploiting the LSTM architecture to encode context, 
we found that we were able to achieve competitive results using only three sentence-position features to model parser state: 
the head word of each of the top two trees on the stack ($s_0$ and $s_1$), and the next word on the queue ($q_0$); see Table~\ref{tab:features}. 

The usefulness of the head words on the stack is clear enough, since those are the two words that are linked by a dependency when taking a reduce action. The next incoming word on the queue is also important because the top tree on the stack should not be reduced if it still has children which have not yet been shifted. That feature thus allows the model to learn to delay a right-reduce until the top tree on the stack is fully formed, shifting instead.

\subsection{Hierarchical Classification}

The structure of our network model after computing positional features is fairly straightforward and similar to previous neural-network parsing approaches such as \namecite{chen2014fast} and \namecite{weiss2015google}.
It consists of a multilayer perceptron using a single ReLU hidden layer followed by a linear classifier over the action space, with the training objective being negative log softmax. 

We found that performance could be improved, however, by factoring out the decision over structural actions (i.e., shift, left-reduce, or right-reduce) and the decision of which arc label to assign upon a reduce.
We therefore use separate classifiers for those decisions, each with its own fully-connected hidden and output layers but sharing the underlying recurrent architecture. This structure was used for the results reported in Section~\ref{sec:exps}, and it is referred to as ``Hierarchical Actions" when compared against a single action classifier in Table~\ref{table:ablation}.


\section{Shift-Promote-Adjoin Constituency~Parsing}
\label{sec:shift-promote-adjoin}
\begin{figure}
\centering
\hspace{-0.15in}
  \begin{tabular}{ll}
   input:      & $w_0 \ldots w_{n-1}$ \\[0.06in]
    axiom        & \olditem{0}{0}{\epsilon}: $\emptyset$ \\ [0.06in]

    \small{\shift}       & \inferrule{\olditem{\ell}{j}{S}}{\olditem{\ell+1}{j+1}{S\mid j}} \ $j<n$ \\[0.2in]

    \small{$\promote(X)$}     & \inferrule{\olditem{\ell}{j}{S\mid t}}
                             {\olditem{\ell+1}{j}{S \mid X(t)}} \\[0.2in]

    \small{\ladj}     & \inferrule{\olditem{\ell}{j}{S\mid  t \mid  X (t_1 ... t_k)}}
                             {\olditem{\ell+1}{j}{S\mid X (t, t_1 ... t_k)}} \\[0.2in]


    goal         & \olditem{2n-1}{n}{\sttop} 
  \end{tabular}
\caption{Our shift-promote-adjoin system
for constituency parsing (\radj omitted).
}
\label{fig:vanilla}
\end{figure}

\newcommand{\wordtag}[3]{\raisebox{-.2cm}{\begin{tabular}{c}\!\!\!$^{#3}${#1}\\[-0.05cm]\small {\ \ #2}\end{tabular}}}

\begin{figure}
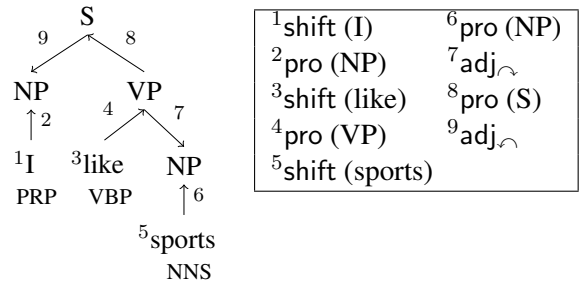

\centering
\tikzset{sibling distance=-0.4cm}
\hspace{-0.5cm}
\resizebox{3.5cm}{!}{
\Tree[.S \edge[->] node[auto=right] {$_9$}; [.NP \edge[<-] node[auto=left]{$^2$}; \wordtag{I}{PRP}{1} ]  
       \edge[<-] node[auto=left]{$_8$}; 
            [.VP  \edge[<-] node[auto=right] {$_4$}; \wordtag{like}{VBP}{3} 
               \edge[->] node[auto=left] {$_7$}; 
               [.NP  \edge[<-] node[auto=left]{$^6$}; \wordtag{sports}{NNS}{5} ] ] 
     ]
}
\hspace{-0.01cm}\raisebox{-1.1cm}{
\begin{tabular}{|p{2.2cm}p{1.2cm}|}
\hline
$^1$\shift(I)   & \hspace{-0.4cm} $^6$\promote(NP)\\
$^2$\promote(NP) & \hspace{-0.4cm}  $^7$\radj\\
$^3$\shift(like)   & \hspace{-0.4cm} $^8$\promote(S) \\
$^4$\promote(VP)  &   \hspace{-0.4cm} $^9$\ladj \\
$^5$\shift(sports) &  \\
\hline
\end{tabular}
}
\caption{Shift-Promote-Adjoin parsing example.
Upward and downward arrows indicate promote and (sister-)adjunction actions, respectively.  
}
\label{fig:spa}
\end{figure}

To further demonstrate the advantage of our idea of minimal features with bidirectional sentence representations, we extend our work from dependency parsing to constituency parsing.
However, the latter is significantly more challenging than the former
under the shift-reduce paradigm
because:
\begin{itemize}
\item we also need to predict the nonterminal labels
\item the tree is not binarized (with many unary rules
  and more than binary branching rules)
\end{itemize}

While most previous work binarizes the constituency tree in 
a preprocessing step \cite{zhu+:2013,wang+xue:2014,mi+huang:2015},
we propose a novel ``Shift-Promote-Adjoin'' paradigm
which does not require any binariziation or transformation of constituency trees
(see Figure~\ref{fig:spa}).
Note in particular that, in our case
only the Promote action produces a new tree node (with a non-terminal label),
while the Adjoin action is the linguistically-motivated ``sister-adjunction'' operation, 
i.e., attachment \cite{chiang:2000,henderson:2003}.
By comparison, in previous work, both Unary-X and Reduce-L/R-X actions 
produce new labeled nodes (some of which are auxiliary nodes due to binarization).
Thus our paradigm has two advantages:
\begin{itemize}
\item it dramatically reduces the number of possible actions, from $3X+1$ or more in previous work 
to $3+X$, where $X$ is the number of nonterminal labels,
which we argue would simplify learning;
\item it does not require binarization \cite{zhu+:2013,wang+xue:2014} or compression of unary chains \cite{mi+huang:2015}
\end{itemize}

There is, however, a more closely-related ``shift-project-attach'' paradigm 
by \namecite{henderson:2003}.
For the example in Figure~\ref{fig:spa} he would use the following actions:
\setlength{\exampleindent}{15pt}
\begin{examples}
\item[]
{shift(I), project(NP), project(S), shift(like),
project(VP), shift(sports), project(NP), attach, attach.}
\end{examples}
The differences are twofold:
first, our Promote action is head-driven, 
which means we only promote the head
child (e.g., VP to S) whereas his Project action promotes the {\em first} child (e.g., NP to S);
and secondly, as a result, his Attach action is always right-attach whereas our Adjoin action could be 
either left or right.
The advantage of our method is its close resemblance to shift-reduce dependency parsing,
which means that our constituency parser is jointly 
performing both tasks and can produce both kinds of trees. This also means that we use head rules to determine the correct order of gold actions.

We found that in this setting, we did need slightly more input features. As mentioned, node labels are necessary to distinguish whether a tree has been sufficiently promoted, and are helpful in any case. We used 8 labels: the current and immediate predecessor label of each of the top two stacks on the tree, as well as the label of the left- and rightmost adjoined child for each tree. We also found it helped to add positional features for the leftmost word in the span for each of those trees, bringing the total number of positional features to five. See Table~\ref{tab:features} for details.



\section{Experimental Results}
\label{sec:exps}
We report both dependency and constituency parsing results 
on both English and Chinese.

All experiments were conducted with minimal hyperparameter tuning. The settings used for the reported results are summarized in Table~\ref{tab:settings}. Networks parameters were updated using gradient backpropagation, including backpropagation through time for the recurrent components, using ADADELTA for learning rate scheduling \cite{Zeiler2012}. We also applied dropout \cite{Hinton2012} (with $p=0.5$) to the output of each LSTM layer (separately for each connection in the case of the two-layer network).

We tested both types of parser on the Penn Treebank (PTB) and Penn Chinese Treebank (CTB-5), with the standard splits for each of training, development, and test sets. Automatically predicted part of speech tags with 10-way jackknifing were used as inputs for all tasks except for Chinese dependency parsing, where we used gold tags, following the traditions in literature.

\subsection{Dependency Parsing: English \& Chinese}

Table~\ref{table:english_dep_results} shows results for English Penn Treebank using Stanford dependencies. Despite the minimally designed feature representation, relatively few training iterations, and lack of pre-computed embeddings, the parser performed on par with state-of-the-art incremental dependency parsers,
and slightly outperformed the state-of-the-art greedy parser.

The ablation experiments shown in the Table~\ref{table:ablation} indicate 
that both forward and backward contexts for each word are very important
to obtain strong results. 
Using only word forms and no part-of-speech input similarly degraded performance.

\begin{table}[H]
\centering
\resizebox{.48\textwidth}{!}{
\begin{tabular}{| l | p{.7cm}p{.75cm} | p{.7cm}p{.75cm} |}
   \hline
   \multirow{2}{*}{Parser}  &  \multicolumn{2}{ c |}{Dev} & \multicolumn{2}{ c |}{Test} \\
                            & UAS & LAS & UAS & LAS \\
   \hline
   C \& M 2014 & 92.0 & 89.7 & 91.8 & 89.6 \\
   Dyer et al.~2015 & 93.2 & 90.9 & 93.1 & 90.9 \\
   Weiss et al.~2015 & - & - & 93.19 & 91.18 \\
   + Percept./Beam & - & - & 93.99 & 92.05\\
   \hline
   Bi-LSTM         & 93.31 & 91.01 & 93.21 & 91.16 \\
   2-Layer Bi-LSTM & 93.67 & 91.48 & 93.42 & 91.36 \\
    \hline
\end{tabular}
}
\caption{Development and test set results for shift-reduce dependency parser on Penn Treebank using only ($s_1$, $s_0$, $q_0$) positional features.}
\label{table:english_dep_results}
\end{table}


\begin{table}[H]
\centering
\begin{tabular}{| l | ll |}
   \hline
   Parser & UAS & LAS \\
   \hline
   Bi-LSTM Hierarchical$^\dagger$   & 93.31 & 91.01 \\ 
   $\dagger$ - Hierarchical Actions & 92.94 & 90.96 \\
   $\dagger$ - Backward-LSTM        & 91.12 & 88.72 \\
   $\dagger$ - Forward-LSTM         & 91.85 & 88.39 \\
   $\dagger$ - tag embeddings       & 92.46 & 89.81 \\
    \hline
\end{tabular}
\caption{Ablation studies on PTB dev set (wsj 22).  
Forward and backward context, 
and part-of-speech input were all critical to strong performace.}
\label{table:ablation}
\end{table}

\begin{figure}[ht]
\centering
\includegraphics[width=0.45\textwidth]{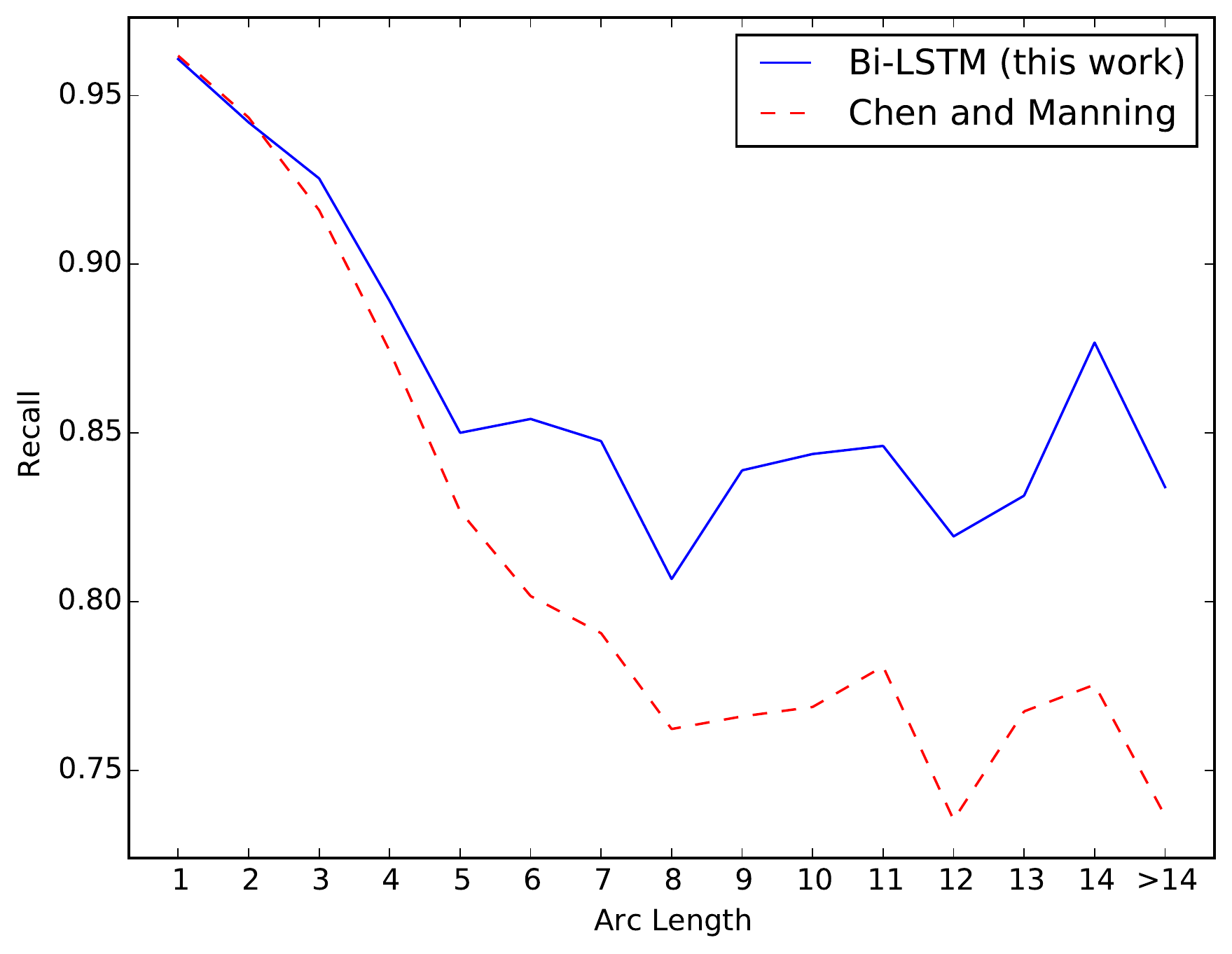}
\caption{Recall on dependency arcs of various lengths in PTB dev set. The Bi-LSTM parser is particularly good at predicting longer arcs.}
\label{fig:arc_recall}
\end{figure}

Figure~\ref{fig:arc_recall} compares our parser with that of \namecite{chen2014fast} 
in terms of arc recall for various arc lengths. 
While the two parsers perform similarly on short arcs,
ours significantly outpeforms theirs on longer arcs,
and more interestingly our accuracy does not degrade much after length 6.
This confirms the benefit of having a global sentence repesentation in our model.

Table~\ref{tab:chndep} summarizes the Chinese dependency parsing results.
Again, our work is competitive with the state-of-the-art greedy parsers.

\begin{table}[H]
\centering
\resizebox{.48\textwidth}{!}{
\begin{tabular}{| l | p{.7cm}p{.75cm} | p{.7cm}p{.75cm} |}
   \hline
   \multirow{2}{*}{Parser}  &  \multicolumn{2}{ c |}{Dev} & \multicolumn{2}{ c |}{Test} \\
                            & UAS & LAS & UAS & LAS \\
   \hline
   C \& M 2014 & 84.0 & 82.4 & 83.9 & 82.4  \\
   Dyer et al.~2015 & 87.2 & 85.9 & 87.2 & 85.7 \\

   \hline
   Bi-LSTM         & 85.84 & 85.24 & 85.53 & 84.89 \\
   2-Layer Bi-LSTM & 86.13 & 85.51 & 86.35 & 85.71 \\

    \hline
\end{tabular}
}
\caption{Development and test set results for shift-reduce dependency parser on Penn Chinese Treebank (CTB-5) using only ($s_1$, $s_0$, $q_0$) position features (trained and tested with gold POS tags).}
\label{tab:chndep}
\end{table}

\subsection{Constituency Parsing: English \& Chinese}

Table~\ref{tab:chnconst}
compares our constituency parsing results 
with state-of-the-art incremental parsers.
Although our work are definitely less accurate than 
those beam-search parsers, 
we achieve the highest accuracy among greedy parsers,
for both English and Chinese.\footnote{
The greedy accuracies for \namecite{mi+huang:2015} are from Haitao Mi, and greedy results for \namecite{zhu+:2013} come from duplicating experiments with code provided by those authors.}$^,$\footnote{
The parser of \namecite{vinyals2015} does not use an explicit transition system,
but is similar in spirit since generating a right bracket can be viewed as a reduce action.}

\begin{table}[H]
\centering
\resizebox{.48\textwidth}{!}{
\begin{tabular}{| l | c | p{.7cm}p{.7cm} | p{.7cm}p{.7cm} |}
   \hline
   \multirow{2}{*}{Parser} & \multirow{2}{*}{$b$}
                              &  \multicolumn{2}{ c |}{English} & \multicolumn{2}{ c |}{Chinese} \\
                            & &\small{greedy} & \small{beam}  & \small{greedy} & \small{beam}  \\
    \hline 
    \namecite{zhu+:2013} & 16 & 86.08 & 90.4 & 75.99 & 85.6 \\
    Mi \& Huang (05)  & 32 &  84.95 & 90.8 & 75.61 & 83.9 \\
    Vinyals et al.~(05) & 10 & - & 90.5 & - & - \\ 
    \hline
    Bi-LSTM              & - & 89.75 & - & 79.44 & - \\
    2-Layer Bi-LSTM      & - & {\bf 89.95} & - & {\bf 80.13} & - \\                        
    \hline
\end{tabular}
}
\caption{Test F-scores for constituency parsing on Penn Treebank and CTB-5.} 
\label{tab:chnconst}
\end{table}

\begin{table}[H]
\centering
\resizebox{.5\textwidth}{!}{
\begin{tabular}{| l | c | c |}
    \hline
    & Dependency & Constituency \\
    \hline
    \multicolumn{3}{| l |}{\bf Embeddings} \\
    \hline
    Word (dims) & 50 & 100 \\
    Tags (dims) & 20 & 100 \\
    Nonterminals (dims) & - & 100 \\
    Pretrained & No & No \\

    \hline
    \multicolumn{3}{| l |}{\bf Network details} \\
    \hline
    LSTM units (each direction) & 200 & 200 \\
    ReLU hidden units      & 200 / decision & 1000 \\    

    \hline
    \multicolumn{3}{| l |}{\bf Training} \\
    \hline
    Training epochs & 10 & 10 \\
    Minibatch size (sentences) & 10 & 10 \\
    Dropout (LSTM output only) & 0.5 & 0.5 \\
    L2 penalty (all weights) & none & $1 \times 10^{-8}$ \\
    ADADELTA $\rho$ & 0.99 & 0.99 \\
    ADADELTA $\epsilon$      & $1 \times 10^{-7}$ & $1 \times 10^{-7}$ \\ 
    \hline
\end{tabular}
}
\caption{Hyperparameters and training settings.} 
\label{tab:settings}
\end{table}

\section{Related Work}
\label{sec:related}

Because recurrent networks are such a natural fit for modeling languages (given the sequential nature of the latter), bi-directional LSTM networks are becoming increasingly common in all sorts of linguistic tasks, for example event detection in \namecite{ghaeini2016}. In fact, we discovered after submission that \namecite{kiperwasser2016} have concurrently developed 
an extremely similar approach to our dependency parser.
Instead of extending it to constituency parsing,
they also apply the same idea to graph-based dependency parsing.

\section{Conclusions}

We have presented a simple bi-directional LSTM sentence representation model
for minimal features in both incremental dependency and incremental constituency parsing,
the latter using a novel shift-promote-adjoin algorithm.
Experiments show that our method are competitive with
the state-of-the-art greedy parsers on both parsing tasks and on both English and Chinese.

\section*{Acknowledgments}
We thank the anonymous reviewers for comments.
We also thank Taro Watanabe, Muhua Zhu, and Yue Zhang for sharing their code,
Haitao Mi for producing greedy results from his parser,
and Ashish Vaswani and Yoav Goldberg for discussions.
The authors were supported in part by DARPA FA8750-13-2-0041 (DEFT), NSF IIS-1449278, 
and a Google Faculty Research Award.


\bibliographystyle{acl}
\bibliography{thesis}

\begin{thebibliography}{}

\bibitem[\protect\citename{Chen and Manning}2014]{chen2014fast}
Danqi Chen and Christopher~D Manning.
\newblock 2014.
\newblock A fast and accurate dependency parser using neural networks.
\newblock In {\em Empirical Methods in Natural Language Processing (EMNLP)}.

\bibitem[\protect\citename{Chiang}2000]{chiang:2000}
David Chiang.
\newblock 2000.
\newblock Statistical parsing with an automatically-extracted tree-adjoining
  grammar.
\newblock In {\em Proc. of ACL}.

\bibitem[\protect\citename{Dyer \bgroup et al.\egroup
  }2015]{dyer2015transition}
Chris Dyer, Miguel Ballesteros, Wang Ling, Austin Matthews, and Noah~A Smith.
\newblock 2015.
\newblock Transition-based dependency parsing with stack long short-term
  memory.
\newblock {\em arXiv preprint arXiv:1505.08075}.

\bibitem[\protect\citename{Ghaeini \bgroup et al.\egroup }2016]{ghaeini2016}
Reza Ghaeini, Xiaoli~Z. Fern, Liang Huang, and Prasad Tadepalli.
\newblock 2016.
\newblock Event nugget detection with forward-backward recurrent neural
  networks.
\newblock In {\em Proc. of ACL}.

\bibitem[\protect\citename{Henderson}2003]{henderson:2003}
James Henderson.
\newblock 2003.
\newblock Inducing history representations for broad coverage statistical
  parsing.
\newblock In {\em Proceedings of NAACL}.

\bibitem[\protect\citename{Hinton \bgroup et al.\egroup }2012]{Hinton2012}
Geoffrey~E. Hinton, Nitish Srivastava, Alex Krizhevsky, Ilya Sutskever, and
  Ruslan Salakhutdinov.
\newblock 2012.
\newblock Improving neural networks by preventing co-adaptation of feature
  detectors.
\newblock {\em CoRR}, abs/1207.0580.

\bibitem[\protect\citename{Hochreiter and Schmidhuber}1997]{hochreiter1997long}
Sepp Hochreiter and J{\"u}rgen Schmidhuber.
\newblock 1997.
\newblock Long short-term memory.
\newblock {\em Neural computation}, 9(8):1735--1780.

\bibitem[\protect\citename{Kiperwasser and Goldberg}2016]{kiperwasser2016}
Eliyahu Kiperwasser and Yoav Goldberg.
\newblock 2016.
\newblock Simple and accurate dependency parsing using bidirectional {LSTM}
  feature representations.
\newblock {\em CoRR}, abs/1603.04351.

\bibitem[\protect\citename{Mi and Huang}2015]{mi+huang:2015}
Haitao Mi and Liang Huang.
\newblock 2015.
\newblock Shift-reduce constituency parsing with dynamic programming and pos
  tag lattice.
\newblock In {\em Proceedings of the 2015 Conference of the North American
  Chapter of the Association for Computational Linguistics: Human Language
  Technologies}.

\bibitem[\protect\citename{Nivre \bgroup et al.\egroup }2006]{nivre+:2006}
Joakim Nivre, Johan Hall, and Jens Nilsson.
\newblock 2006.
\newblock Maltparser: A data-driven parser-generator for dependency parsing.
\newblock In {\em Proc.~of LREC}.

\bibitem[\protect\citename{Nivre}2008]{nivre:2008}
Joakim Nivre.
\newblock 2008.
\newblock Algorithms for deterministic incremental dependency parsing.
\newblock {\em Computational Linguistics}, 34(4):513--553.

\bibitem[\protect\citename{Vinyals \bgroup et al.\egroup }2015]{vinyals2015}
Oriol Vinyals, {\L}ukasz Kaiser, Terry Koo, Slav Petrov, Ilya Sutskever, and
  Geoffrey Hinton.
\newblock 2015.
\newblock Grammar as a foreign language.
\newblock In {\em Advances in Neural Information Processing Systems}, pages
  2755--2763.

\bibitem[\protect\citename{Wang and Xue}2014]{wang+xue:2014}
Zhiguo Wang and Nianwen Xue.
\newblock 2014.
\newblock Joint pos tagging and transition-based constituent parsing in chinese
  with non-local features.
\newblock In {\em Proceedings of ACL}.

\bibitem[\protect\citename{Weiss \bgroup et al.\egroup }2015]{weiss2015google}
David Weiss, Chris Alberti, Michael Collins, and Slav Petrov.
\newblock 2015.
\newblock Structured training for neural network transition-based parsing.
\newblock In {\em Proceedings of ACL}.

\bibitem[\protect\citename{Zeiler}2012]{Zeiler2012}
Matthew~D. Zeiler.
\newblock 2012.
\newblock {ADADELTA:} an adaptive learning rate method.
\newblock {\em CoRR}, abs/1212.5701.

\bibitem[\protect\citename{Zhang and Nivre}2011]{zhang+nivre:2011}
Yue Zhang and Joakim Nivre.
\newblock 2011.
\newblock Transition-based dependency parsing with rich non-local features.
\newblock In {\em Proceedings of ACL}, pages 188--193.

\bibitem[\protect\citename{Zhu \bgroup et al.\egroup }2013]{zhu+:2013}
Muhua Zhu, Yue Zhang, Wenliang Chen, Min Zhang, and Jingbo Zhu.
\newblock 2013.
\newblock Fast and accurate shift-reduce constituent parsing.
\newblock In {\em Proceedings of ACL 2013}.

\end{thebibliography}

\end{document}